\documentclass[10pt,twocolumn,letterpaper]{article}

\usepackage{wacv}
\usepackage{times}
\usepackage{epsfig}
\usepackage{graphicx}
\usepackage{amsmath}
\usepackage{amssymb}
\usepackage{multirow}
\usepackage{colortbl}
\usepackage{booktabs}
\usepackage{siunitx}
\usepackage{makecell}
\usepackage{epstopdf}
\usepackage{fancyhdr}
\usepackage{subcaption}
\usepackage[normalem]{ulem}
\usepackage{xcolor}
\usepackage{xfrac}

\wacvfinalcopy 
\def\wacvPaperID{278} 

\usepackage{soul,xcolor}
\setstcolor{red}

\ifwacvfinal\fi
\begin{document}

\title{Semi-Coupled Two-Stream Fusion ConvNets\\
	for Action Recognition at Extremely Low Resolutions}

\author{Jiawei Chen, Jonathan Wu, Janusz Konrad, Prakash Ishwar \\
Department of Electrical and Computer Engineering, Boston University, Boston, MA 02215\\
{\tt\small \{garychen, jonwu, jkonrad, pi\}@bu.edu}}

\maketitle
\ifwacvfinal\thispagestyle{empty}\fi

\begin{abstract}
Deep convolutional neural networks (ConvNets) have been recently shown to attain state-of-the-art performance for action recognition on standard-resolution videos.
However, less attention has been paid to recognition performance at extremely low resolutions (eLR) (e.g., $16\times12$ pixels).
%
%
%
Reliable action recognition using eLR cameras would address privacy concerns in various application environments such as private homes, hospitals, nursing/rehabilitation facilities, etc.
%
In this paper, we propose a semi-coupled, filter-sharing network that leverages high-resolution (HR) videos during training in order to assist an eLR ConvNet.
%
We also study methods for fusing spatial and temporal ConvNets customized for eLR videos in order to take advantage of appearance and motion information. 
%
%
Our method outperforms state-of-the-art methods at extremely low resolutions on IXMAS ($93.7\%$) and HMDB ($29.2\%$) datasets.
%

\end{abstract}

\section{Introduction}
Human action and gesture recognition has received significant
attention in computer vision and signal processing communities
\cite{Simonyan14c,wang2013action,xia2012view}.
%
%
{Recently, various ConvNet models have been applied in this context
  and achieved substantial performance gains over traditional methods}
that are based on hand-crafted features
\cite{krizhevsky2012imagenet,sharif2014cnn}.
%
Further improvements in the performance have been realized by using a
two-stream ConvNet architecture \cite{simonyan2014two} in which a
spatial network concentrates on learning appearance features from RGB
images while a temporal network takes optical flow snippets as input to
learn dynamics.
%
%
The final decision is made by averaging outputs of the two networks. 
More recent work \cite{ feichtenhofer2016convolutional,lin2015bilinear,park2016combining} suggests fusion of spatial and temporal cues at an earlier stage so the appearance features are registered with motion features before the final decision. 
Results indicate that this approach improves action recognition performance.
\begin{figure}[!ht]
	\centering{\includegraphics[width =1\linewidth]{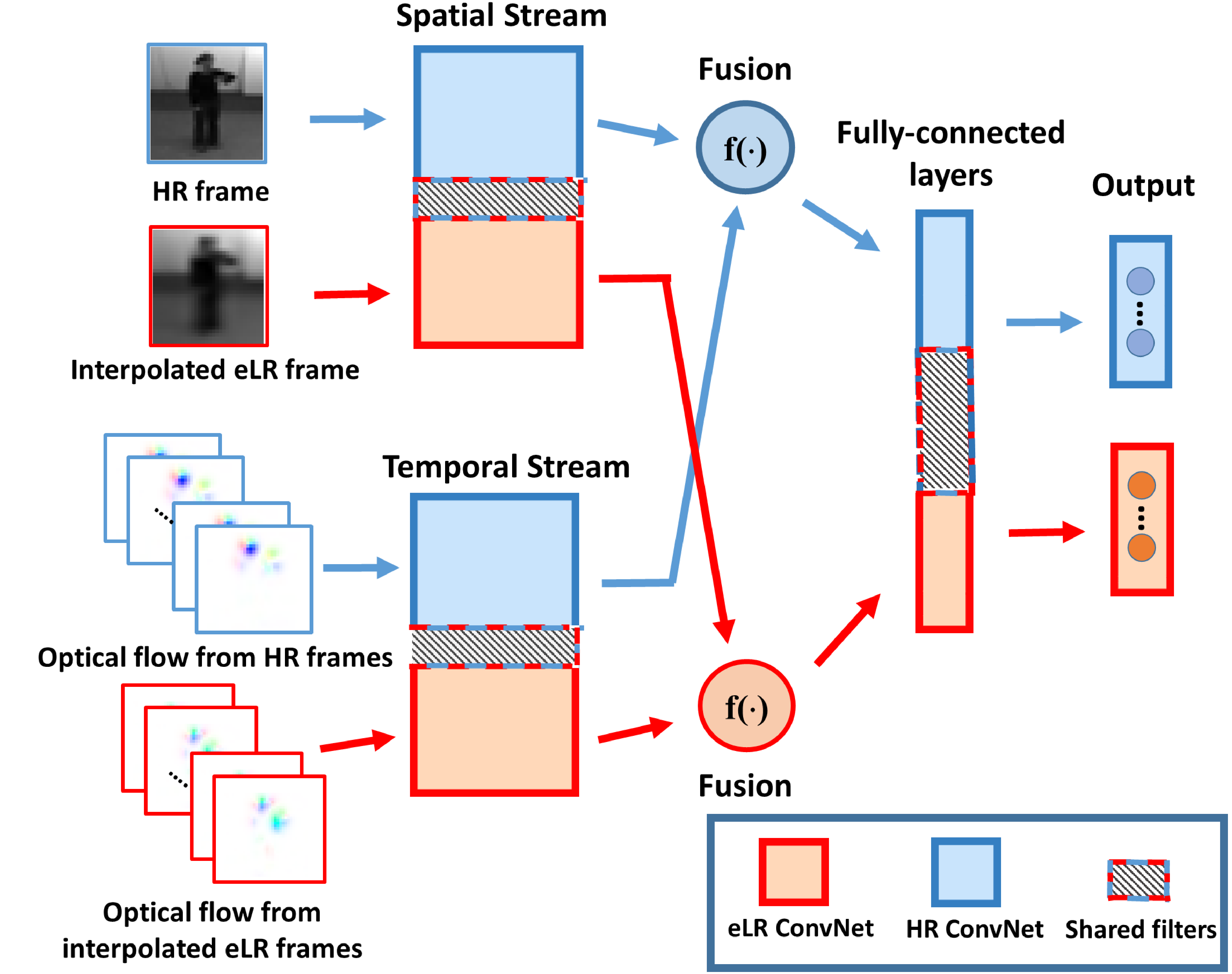}}
	\medskip
	\caption{Diagram of the proposed ConvNet architecture. The
          blue blocks represent training-only HR information. The red
          blocks represent {\it both} training and testing eLR
          information. The overlap between the red and blue blocks
          indicates shared filters between eLR and HR
          information. Please note that both blocks use RGB and
          optical flow information. }
	\vglue -0.3cm
	\label{fig:outline}
\end{figure}

As promising as these recent ConvNet-based models are, they typically rely
upon data at about 200$\times$200-pixel resolution that is likely to
reveal an individual's identity.
%
However, as more and more sensors are being deployed in our homes and offices, the concern for privacy only grows. Clearly, reliable methods for human activity analysis at privacy-preserving resolutions are urgently needed \cite{chen2016estimating, roeper2016}
Some of the early approaches to action recognition from eLR data use simple machine learning algorithms (e.g., nearest-neighbor classifier) \cite{dai2015towards} or leverage a ConvNet but only as an appearance feature extractor \cite{ryoo2016privacy}.
%
%
In very recent work \cite{wang2016studying}, a partially-coupled
super-resolution network (PCSRN) has been proposed for eLR {\it image}
classification (not video).
Basically, this network includes two ConvNets sharing a number of filters
at each convolutional layer.
%
While the input to one ConvNet consists of eLR images only, the input to the other network is formed from the corresponding HR images.
The shared filters are trained to learn a nonlinear mapping between the eLR and HR feature spaces.
%
Although this model has been designed for image recognition tasks, its
excellent performance suggests that filter sharing could perhaps also
benefit action recognition (from video) at extremely low resolutions.

In this paper, we combine the ideas of eLR-HR coupling and of two-stream ConvNets to perform reliable action recognition at extremely low resolutions.
%
{In particular, we adapt an existing end-to-end two-stream fusion
  ConvNet to eLR action recognition.}
%
We provide an in-depth analysis of three fusion methods for spatio-temporal networks, and compare them experimentally on eLR video datasets.
%
%
Furthermore, inspired by the PCSRN model, we propose a semi-coupled two-stream fusion ConvNet that leverages HR videos during training in order to help the eLR ConvNet obtain enhanced discriminative power by sharing filters between eLR and HR ConvNets (Fig.\ref{fig:outline}).
%
Tested on two public datasets, the proposed model outperforms state-of-the-art eLR action recognition methods thus justifying our approach.
	%
	%
	%
	%

\section{Related Work}
ConvNets have been recently applied to action recognition and quickly
yielded state-of-the-art performance.
%
In the quest for further gains, a key question is how to properly
incorporate appearance and motion information in a ConvNet
architecture.
%
%
In \cite{ji20133d,karpathy2014large,tran2014c3d}, various 3D ConvNets were proposed to learn spatio-temporal features by stacking consecutive RGB frames in the input.
In \cite{simonyan2014two}, a novel two-stream ConvNet architecture was proposed which learns two separate networks: one dedicated to spatial RGB information, and another dedicated to temporal optical flow information.
%
The softmax outputs of these two networks are later combined together to provide a final ``joint'' decision.
%
Following this pivotal work, many works have extended the two-stream architecture such that only a single, combined network is trained.
In \cite{lin2015bilinear}, bilinear fusion was proposed in which the last convolutional layers of both networks are combined using an outer-product and pooling.
%
%
Similarly, in \cite{park2016combining} multiplicative fusion was proposed, and in \cite{feichtenhofer2016convolutional} 3D convolutional fusion was introduced (incorporating an additional temporal dimension).
%
%
However, all these methods were applied to standard-resolution video,
and have not, to the best of our knowledge, been applied in the eLR
context.
%
%
%
%

There have been few works that have addressed eLR in the context of visual recognition.
%
In \cite{wang2016studying}, very low resolution networks were
investigated in the context of eLR {\it image} recognition. 
%
The authors proposed to incorporate HR images in training to augment the learning process of the network through filter sharing (PCSRN).
%
%
In \cite{dai2015towards}, eLR action recognition was first explored using $l_1$ nearest-neighbor classifiers to discriminate between action sequences.
%
More recently, egocentric eLR activity recognition was explored in
\cite{ryoo2016privacy}. 
%
%
The authors introduced inverse super resolution (ISR) to learn an optimal set of image transformations during training that generate multiple eLR videos from a single HR video. Then, they trained a classifier based on features extracted from all generated eLR videos. The per-frame features include histogram of pixel intensities, histogram of oriented gradients (HOG) \cite{dalal2005histograms}, histogram of optical flow (HOF)  \cite{chaudhry2009histograms} and ConvNet features. To capture temporal changes, they used the Pooled Time Series (POT) feature representation \cite{ryoo2015pooled} which is based on time-series analysis. This classifier was finally used in testing.
%
%
%
However, in keeping with recent research trends our aim is to develop an end-to-end, ConvNet-only solution that avoids hand-crafted features and, therefore, minimizes human intervention. 
%
We benchmark our proposed methodologies against last two works,
and show consistent recognition improvement.
%
%
%
%
%
%
%
\begin{figure*}[!t]
	\centering{\includegraphics[width=0.95\linewidth]{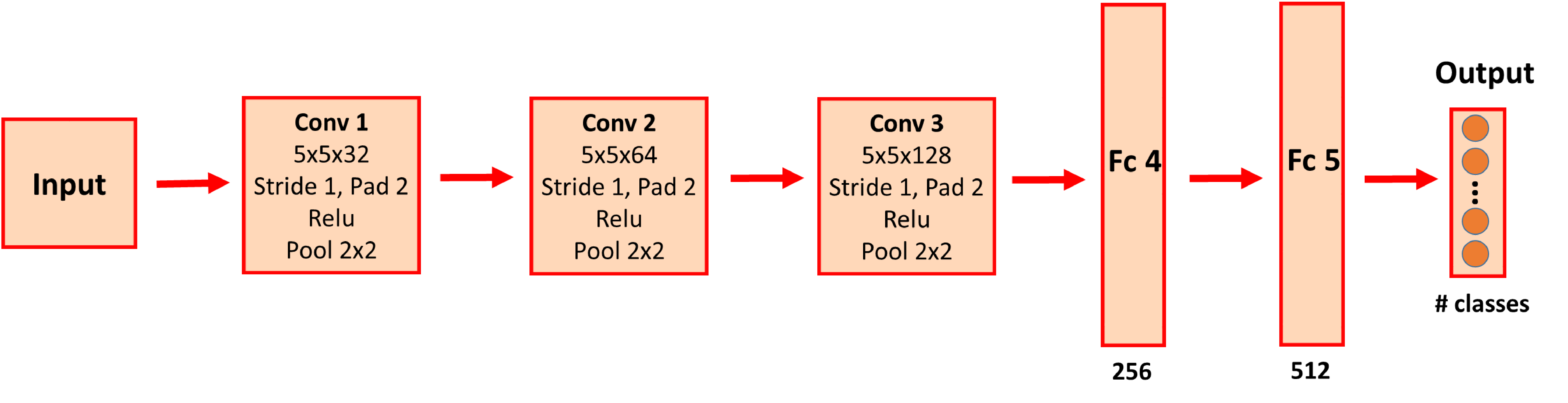}}
	\medskip
	\caption{Basic ConvNet used in our model. The spatial and temporal streams have the same architecture except that the input dimension is larger in the temporal stream (the input to the temporal stream is a stacked optical flow). In our two-stream fusion ConvNets, two base ConvNets are fused after either the ``Conv3'' or ``Fc4'' layer.\label{fig:base_network}}
	\end{figure*}

\section{Technical Approach}
%
%
%
%
%
%
In this section, we propose two improvements to the two-stream architecture in the context of eLR.
First, we explore methods to fuse the spatial and temporal networks, which allows subsequent layers to amplify and leverage joint spatial and temporal features.
Second, we propose using semi-coupled networks which leverage HR information in training to learn transferable features between HR and eLR frames, resembling domain adaptation, in both the spatial and temporal streams.
%
%
\subsection{Fusion of the two-stream networks}
%
%
%
	%
	%
%
%
%
%
%
%
%
Multiple works have extended two-stream ConvNets by combining the spatial and temporal cues such that only a single, combined network is trained \cite{feichtenhofer2016convolutional,lin2015bilinear, park2016combining}.
This is most frequently done by fusing the outputs of the spatial and
temporal network's convolution layers with the purpose of learning a correspondence of activations at the same pixel location. 
In this section, we discuss three fusion methods that we explore and implement in the context of eLR.
%

In general, fusion is applied between a spatial ConvNet and a temporal ConvNet. 
A fusion function $f$:
$f(\boldsymbol{x}_{s}^{n},\boldsymbol{x}_{t}^{n}) \rightarrow
\boldsymbol{y}^{n}$ fuses spatial features at the output of the $n$-th
layer $\boldsymbol{x}_{s}^{n} \in \mathbb{R}^{H_s^n \times W_s^n
  \times D_s^n}$ and temporal features at the output of the $n$-th
layer $\boldsymbol{x}_{t}^{n} \in \mathbb{R}^{H_t^n \times W_t^n
  \times D_t^n}$ to produce the output features $\boldsymbol{y}^{n}
\in \mathbb{R}^{H_{o}^n \times W_{o}^n \times D_{o}^n}$, where $H$,
$W$, and $D$ represent the height, width and the number of channels respectively. 
%
For simplicity, we assume $H_o=H_s=H_t, W_o=W_s=W_t$, and $D_s = D_t$ ($D_o$ is defined below).
%
We discuss the fusion function for three possible operators:

\noindent{\bf Sum Fusion:} Perhaps the simplest fusion strategy is to
compute the summation of two feature maps at the same pixel location
$(i,j)$ and the same channel $d$:
%
\begin{equation}
\boldsymbol{y}^{n,sum}(i,j,d) = \boldsymbol{x}_{s}^n(i,j,d) +
\boldsymbol{x}_{t}^n(i,j,d)
\end{equation}
where $ 1\leq i \leq H_o$ , $ 1\leq j \leq W_o$, $1\leq d \leq D_o$
($D_o = D_s = D_t$) and $\boldsymbol{x}_{s}^n$, $\boldsymbol{x}_{t}^n$,
$\boldsymbol{y}^n \in \mathbb{R}^{H_o\times W_o \times D_o}$.
The underlying assumption of summation fusion is that the spatial and
temporal feature maps from the same channel will share similar contexts. 
%

\noindent{\bf Concat Fusion:} The second fusion method we consider is
a concatenation of two feature maps at the same spatial location $(i,j)$
across channel $d$:
\begin{equation}
	\label{eqn:concat}
   \boldsymbol{y}^{n,cat}(i,j,2d) = \boldsymbol{x}_{s}^n(i,j,d), \ 
\end{equation}
\begin{equation}
\label{eqn:concat2}
\boldsymbol{y}^{n,cat}(i,j,2d+1) = \boldsymbol{x}_{t}^n(i,j,d)
\end{equation}
%
where $\boldsymbol{y}^{n,cat} \in \mathbb{R}^{H_o \times W_o\times
  D_o}$, $D_o = D_s + D_t$.
Unlike the summation fusion, the concatenation fusion does not
actually blend the feature maps together.
%

\noindent{\bf Conv Fusion:} The third fusion operator we explore is
convolutional fusion.
First, $\boldsymbol{x}_{s}^{n}$ and $\boldsymbol{x}_{t}^{n}$ are
concatenated as shown in (\ref{eqn:concat}, \ref{eqn:concat2}).
Then, the stacked up feature map is convolved with a bank of filters $\mathcal{F} \in \mathbb{R}^{1\times 1\times D_o \times D'_{o}}$ as follows:
%
\begin{equation}
\boldsymbol{y}^{n,conv} = \boldsymbol{y}^{n,cat} * \mathcal{F} + b,
\end{equation}
where $b\in \mathbb{R}^{D'_o}$ is a bias term. 
The filters have dimensions $1 \times 1 \times D_o$, 
$D_o = D_s + D_t$ and are used to learn weighted combinations of feature maps $\boldsymbol{x}_{s}^{n}, \boldsymbol{x}_{t}^{n}$ at a shared pixel location. 
%
For our experiments, we have set the number of filters to $D'_o = 0.5 \times D_o$. 

%

Note, that regardless of the chosen fusion operator, the network will select filters throughout the entire network so as to minimize loss, and optimize recognition performance.
Also, we would like to point out that other fusion operators, such as max, multiplication, and bilinear fusion \cite{lin2015bilinear}, are possible, but have been shown in \cite{feichtenhofer2016convolutional} to perform slightly worse than the operators we've discussed.
%
%
%
%
Finally, it is worth noting that the type of fusion operation and the
layer in which it occurs have a significant impact on the number of
parameters. The number of parameters can be quite small if fusion
across networks occurs in early layers. 
%
For example, convolutional fusion requires additional parameters since 
introducing a convolutional layer requires more filters.
%
Regarding where to fuse the two networks, we adopt the convention used
in \cite{feichtenhofer2016convolutional} to fuse the two networks
after their last convolutional layer (see
Fig.\ref{fig:semi-coupled-fusion network}). 
%
We later report the results of fusion after the last convolutional
layer (Conv3) and the first fully-connected layer (Fc4) and contrast
their classification performance. 
%

\subsection{Semi-coupled networks} \label{subsec:semi-couple}

Applying recognition directly to eLR video is not robust as visual features tend to carry little information \cite{wang2016studying}. 
%
However, it is possible to augment ConvNet training with an auxiliary, HR version of the eLR video, but only use an eLR video in testing. 
%
%
%
\begin{figure*}[!ht]
	\centering{\includegraphics[width=1\linewidth]{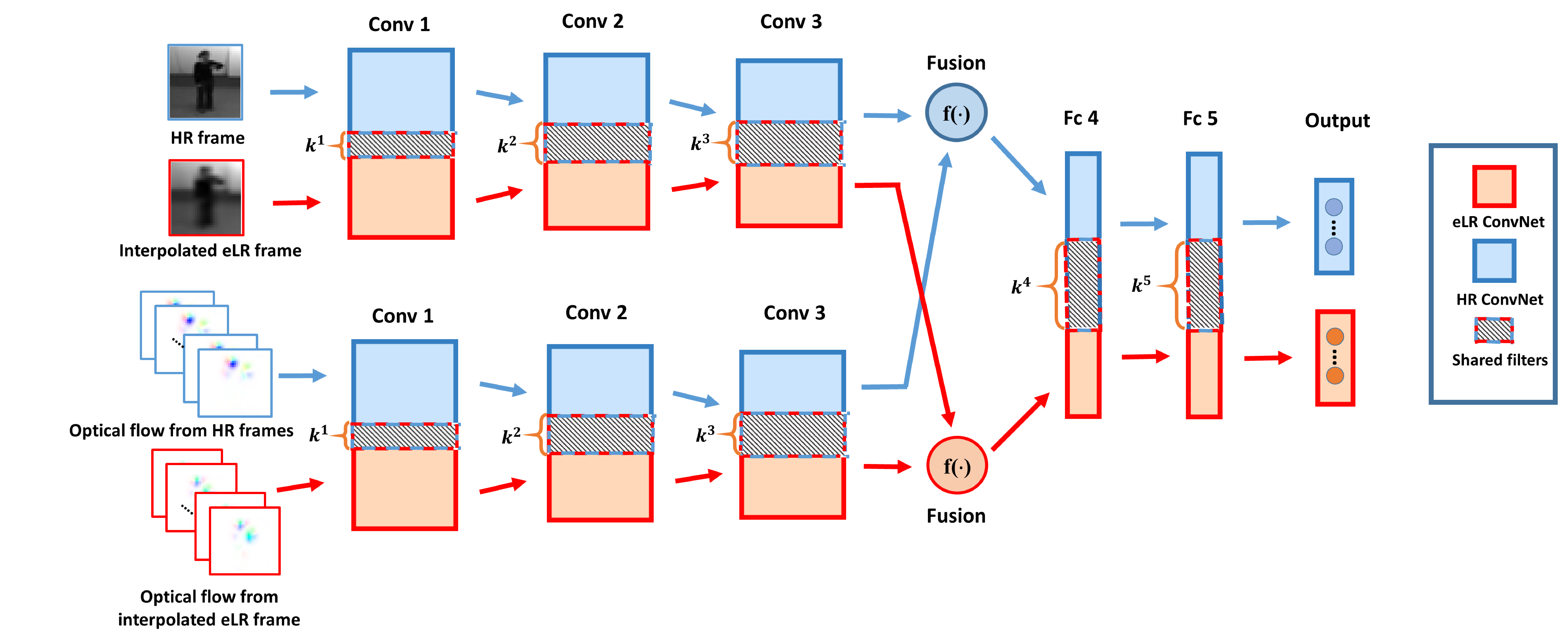}}
	\medskip
	\caption{Visualization of the proposed semi-coupled networks
%
%
	of two fused two-stream ConvNets for video recognition. We
        feed HR RGB and optical flow frames ($32\times32$ pixels) to
        the HR ConvNet (colored in blue). We feed eLR RGB
        ($16\times12$ interpolated to $32\times32$ pixels) and optical
        flow frames (computed from the interpolated $32\times32$ pixel
        RGB frames) to the eLR ConvNet (colored in red). In training,
        the two ConvNets share $k^n$ ($n=1,...,5$) filters (gray
        shaded) between corresponding convolutional and fully-connected layers. Note that the deeper the layer, the
        more filters are being shared. In testing, we decouple the two
        ConvNets and only use the eLR network (the red network which
        includes the shared filters).  } \vglue -0.3cm
	\label{fig:semi-coupled-fusion network}
\end{figure*}
In this context, we propose to use semi-coupled networks which share filters between an eLR and an HR fused two-stream ConvNets.
%
%
The eLR two-stream ConvNet takes an eLR RGB frame and its
corresponding eLR optical flow frames as input.
As we will discuss later, each RGB frame corresponds to multiple optical flow frames.
The eLR RGB frames are interpolated to $32\times32$ pixels from their
original $16\times12$ resolution.
The eLR optical flow is computed from the interpolated $32\times 32$ eLR RGB frames.
%
The HR two-stream ConvNet simply takes HR RGB and its corresponding HR
optical flow frames of size $32\times32$ as input.
%
In layer number $n$ of the network $(n=1,\ldots,5)$, the eLR and HR
two-stream ConvNets share $k^n$ filters.
%
During training, we leverage both eLR and HR information, and update
the filter weights of both networks in tandem.
During testing, we decouple these two networks and only use the eLR
network which includes shared filters.
This entire process is illustrated in
Fig.~\ref{fig:semi-coupled-fusion network}.

The motivation for sharing filters is two-fold: first, sharing
resembles domain adaptation, aiming to learn transferable features
from the source domain (eLR images) to the target domain (HR images);
second, sharing can be viewed as a form of data augmentation with
respect to the original dataset, as the shared filters will see both
low and high resolution images (doubling the number of training
inputs).
However, it is important to note that in practice, as shown in
\cite{lui2009meta}, the mapping between eLR and HR feature space is
difficult to learn.
As a result, the feature space mapping between resolutions may not
fully overlap or correspond properly to one-another after learning.
%
To address this, we intentionally leave a number of filters ($D^n -
k^n$) unshared in layer $n$, for each $n$.
%
These unshared filters will learn domain-specific (resolution
specific) features, while the shared filters learn the nonlinear
transformations between spaces.
To implement this filter sharing paradigm, we alternate between
updating the eLR and HR two-stream ConvNets during training.
%
Let $\boldsymbol{\theta}_{\text{eLR}}$ and
$\boldsymbol{\theta}_{\text{HR}}$ denote the filter weights of the eLR
and the HR two-stream ConvNets.
%
These two filters are composed of three types of weights:
$\boldsymbol{\theta}_{\text{shared}}$, the weights that are shared
between both the eLR and HR networks, and
$\boldsymbol{\theta}_{\text{eLR}^*}$,
$\boldsymbol{\theta}_{\text{HR}^*}$, the weights that belong to only
the eLR or the HR network, respectively.
%
%
%
With these weights, we update both networks as follows:
%
\begin{equation}
\boldsymbol{\theta}_{\text{eLR}}^m = 
\left[\begin{array}{c} \boldsymbol{\theta}_{\text{eLR}^*}^{m-1} + \mu
    \dfrac{\partial \boldsymbol{L}_\text{eLR}^{m-1}}{\partial
      \boldsymbol{\theta}_{\text{eLR}^*}^{m-1} }\\ \\
 \boldsymbol{\theta}_{\text{shared}}^{2m-2} +
\mu
{\dfrac{\partial \boldsymbol{L}_\text{eLR}^{m-1}}{\partial
    \boldsymbol{\theta}_{\text{shared}}^{2m-2}}}
\end{array}\right]
\end{equation}
\begin{equation}
\boldsymbol{\theta}_{\text{HR}}^m = 
\left[\begin{array}{c} \boldsymbol{\theta}_{\text{HR}^*}^{m-1}
+
\mu {\dfrac{\partial \boldsymbol{L}_\text{HR}^{m-1}}{\partial
    \boldsymbol{\theta}_{\text{HR}^*}^{m-1} }} \\ \\
%
\boldsymbol{\theta}_{\text{shared}}^{2m-1}
+  \mu
%
{\dfrac{\partial \boldsymbol{L}_\text{HR}^{m-1}}{\partial
    \boldsymbol{\theta}_{\text{shared}}^{2m-1}}}
\end{array}\right]
\end{equation}
where $\mu$ is the learning rate, $m$ is the training iteration, and
$\boldsymbol{L}_\text{eLR}$ and $ \boldsymbol{L}_\text{HR}$ are,
respectively, the loss functions of each network.
In each training iteration, the shared weights are updated in {\it
  both} the eLR and the HR ConvNet, i.e., they are updated twice in
each iteration. Specifically, in each training iteration $m$, we have
\begin{equation}
\boldsymbol{\theta}_{\text{shared}}^{2m-1} = 
\boldsymbol{\theta}_{\text{shared}}^{2m-2} + \mu {\dfrac{\partial
    \boldsymbol{L}_\text{eLR}^{m-1}}{\partial
    \boldsymbol{\theta}_{\text{shared}}^{2m-2}}}
\end{equation}
\begin{equation}
\boldsymbol{\theta}_{\text{shared}}^{2m-2} =
\boldsymbol{\theta}_{\text{shared}}^{2m-3} + \mu {\frac{\partial
    \boldsymbol{L}_\text{HR}^{m-2}}{\partial
    \boldsymbol{\theta}_{\text{shared}}^{2m-3}}}
\end{equation}
However, the resolution-specific unshared weights are only updated
once: either in the eLR ConvNet training update or in the HR ConvNet
training update.
Therefore, the shared weights are updated twice as often as the
unshared weights.

Our approach has been inspired by Partially-Coupled Super-Resolution
Networks (PCSRN) \cite{wang2016studying} where it was shown that
leveraging HR images in training of such networks can help discover
discriminative features in eLR images that would otherwise have been
overlooked during image classification.
PCSRN is a super-resolution network that pre-trains network weights
using filter sharing.
This pre-training is intended to minimize the MSE of the output image
and the target HR image via super-resolution.
%
In our approach, we differ from this work by leveraging HR information
throughout the {\it entire} training process.
Our method does not need to pre-train the network; instead, we learn
the entire network from scratch, and minimize the classification loss
function directly while still incorporating HR information as shown in
the equations above.
Overall, we extend this model in two aspects: first, we consider
shared filters in the fully-connected layers (previously only
convolutional layers were considered for filter sharing).
%
Second, we adapt this method for action recognition in fused
two-stream ConvNets.
We also report results for semi-coupled two-stream ConvNets across
various fusion operators.
\subsection{Implementation details} \label{sec:implementation}
\textbf{Two-stream fusion network}.
%
%
Conventional standard-resolution ConvNet architectures can be
ill-suited for eLR images due to large receptive fields that can
sometimes be larger than the eLR image itself.
%
%
%
%
To address this issue, we have designed an eLR ConvNet consisting of 3 convolutional layers, and 2 fully-connected layers as shown in
Fig.~\ref{fig:base_network}.
We have tried many variations, but found that larger models do not improve performance. Also, the model in \cite{ryoo2016privacy} is larger than ours, but achieves lower CCR.
%
We base both our spatial and temporal streams on this ConvNet, and explore fusion operations after either the ``Conv3'' or ``Fc4'' layer.
%
We train all networks from scratch using the Matconvnet toolbox \cite{vedaldi2015matconvnet}. 
%
The weights are initialized to be zero-mean Gaussian with a small standard deviation of $10^{-3}$.
The learning rate starts from $0.05$ and is reduced by a factor of 10 after every 10 epochs. 
Weight decay and momentum are set to 0.0005 and 0.9 respectively. 
We use a batch size of 256 and perform batch normalization after each convolutional layer. 
%
At every iteration, we perform data augmentation by allowing a 0.5 probability that a given image in a batch is reflected across the vertical axis.
%
Each RGB frame in the spatial stream corresponds to 11 stacked frames of optical flow.
This stacked optical-flow block contains the current, the 5 preceding, and 5 succeeding optical flow frames.
%
To regularize these networks during training, we set the dropout ratio of both fully-connected layers to 0.85.

\textbf{Semi-coupled ConvNets}.
In Section \ref{subsec:semi-couple}, we have discussed how to incorporate filter sharing in a semi-coupled network.
%
%
However, it is not obvious how many filters should be shared in each layer.
To discover the proper proportion of filters we should share, we
conducted a coarse grid search for the coupling ratio $c_n$ from 0 to
1 with a step size of $0.25$.
%
The coupling ratio is defined as:
\begin{equation}
c_n = \frac{k^n}{D^n},\quad n = 1,\cdots,5
\end{equation}
where the two ConvNets are uncoupled when $c_n = 0$ $(n =
1,\cdots,5)$.
%
For the step sizes that we consider, a brute force approach would be
unfeasible, as the total number of two-stream networks to train would
be $5^5=3125$.
%
Therefore, we follow the methodology used in \cite{wang2016studying}
to monotonically increase the coupling ratios with the increasing layer
depth.
This is inspired by the notion that the disparity between eLR and HR
domains is reduced as the layer gets deeper \cite{glorot2011domain,
  wang2014deeply}.
%
For all our experiments, we used the following coupling ratios: $c_1 =
0$, $c_2 = 0.25$, $c_3 = 0.5$, $c_4 = 0.75$, and $c_5 = 1$.
We determined these ratios by performing a coarse grid search on a
cross-validated subset of the IXMAS dataset (subjects 2, 4, 6).
%
%

\textbf{Normalization}. \label{subsection:normalization}
In our experiments, we apply a variant of mean-variance normalization to each video $v_{i,j}[t], i,j = 1,\cdots,R, t = 1,\cdots,T$, where $R$ is the spatial size, $T$ is the temporal length,  and $v_{i,j}[t]$ denotes the grayscale value of pixel $(i,j)$ at time $t$, as follows:
\begin{equation}
\hat{v}_{i,j}[t] = \frac{v_{i,j}[t]-\mu_{i,j}}{\sigma}.
\end{equation}
Above, $\mu_{i,j}$ denotes the empirical mean pixel value across time
for the spatial location $(i,j)$, and $\sigma$ denotes the empirical
standard deviation across all pixels in one video.
The subtraction of the mean emphasizes a subject's local dynamics,
while the division by the empirical standard deviation compensates for
the variability in subject's clothing.

\textbf{Optical flow}.
%
As discussed earlier, we use a stacked block of optical flow frames as
input to the temporal stream.
%
We follow \cite{Wu_2016_CVPR_Workshops} and use colored optical-flow frames.
First, we compute optical flow between two consecutive normalized RGB
frames \cite{liu2009beyond}.
The computed optical flow vectors are then mapped into polar
coordinates and converted to hue and saturation based on the
magnitude and orientation, respectively.
%
The brightness is set to one.
%
As a reminder, the eLR optical flow is computed from the interpolated
$32\times32$ pixel eLR frames.
Further, we subtract the mean of the stacked optical flows to
compensate for global motion as suggested in \cite{simonyan2014two}.

{\bf Source code:} More implementation details as well as source code are available on project web site \cite{code}.
\section{Experiments}
\begin{figure*}[t!]
	\centering
	\begin{subfigure}[t]{0.5\textwidth}
		\centering
		\includegraphics[height=1.9in]{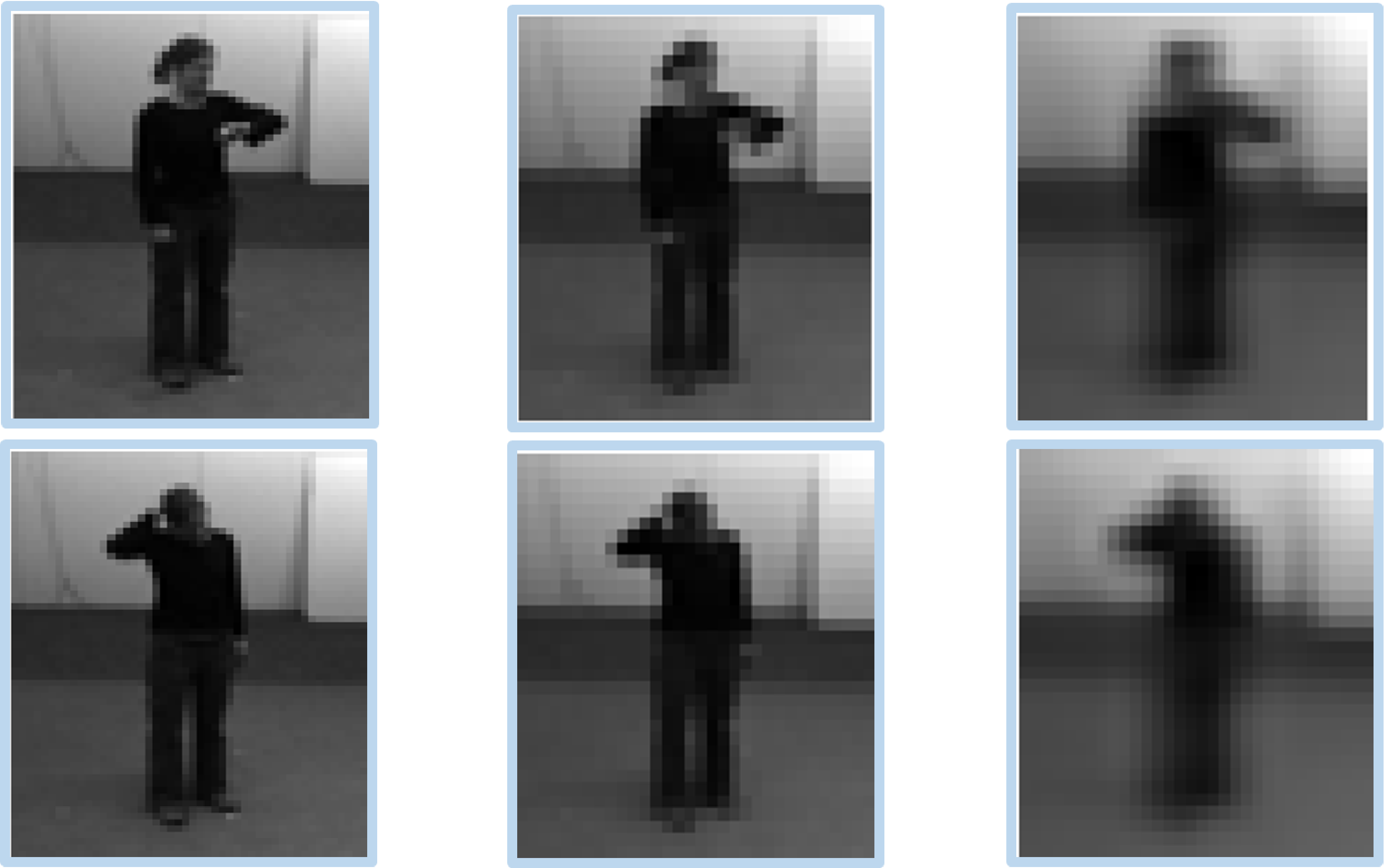}
		\caption{IXMAS}
	\end{subfigure}
	~
	\begin{subfigure}[t]{0.48\textwidth}
		\centering
		\includegraphics[height=1.9in]{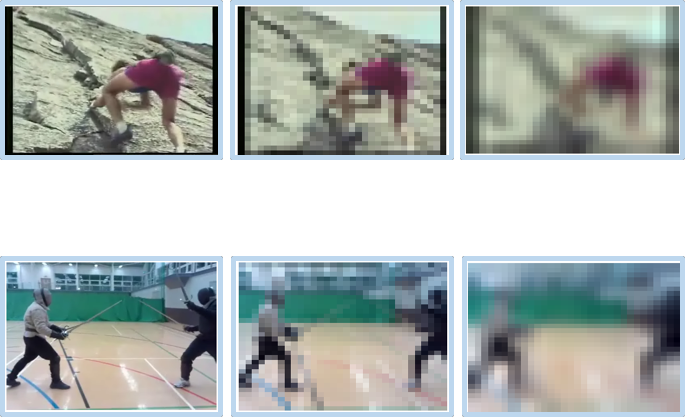}
		\caption{HMDB}
	\end{subfigure}%
	~
	\caption{Sample frames from IXMAS and HMDB datasets. (a) From
          left to right are original frames, and resized $32\times32$
          and $16\times12$ frames from the IXMAS dataset. (b) From
          left to right are original frames, and resized $32\times32$
          and $12\times16$ frames from the HMDB dataset. Note that we
          resize the IXMAS dataset to $16\times12$ and the HMDB
          dataset to $12\times16$ in order to preserve the
          original aspect ratio. We use $32\times32$ resized videos as
          HR data. The $16\times12$ ($12\times16$) eLR videos are
          upscaled using bi-cubic interpolation to $32\times32$
          interpolated-eLR video which is used in our proposed
          semi-coupled fused two-stream ConvNet architecture.}
	\label{fig:dataset}
\end{figure*}
\subsection{Datasets}
In order to confirm the effectiveness of our proposed method, we
conducted experiments on two publicly-available video datasets.
First, we use the ROI sequences from the multi-view IXMAS action dataset, where each subject occupies most of the field of view \cite{weinland10a}.
This dataset includes 5 camera views, 12 daily-life motions each
performed 3 times by 10 actors in an indoor scenario.
Overall, it contains 1,800 videos.
To generate the eLR videos (thus eLR-IXMAS), we decimated the original frames to $16\times12$ pixels and then upscaled them back to $32\times32$ pixels by bi-cubic interpolation (Fig.~\ref{fig:dataset}). 
The upscaling operation does not introduce new information (fundamentally, we are still working with $16\times12$ pixels) but ensures that eLR frames have enough spatial support for hierarchical convolutions to facilitate filter sharing. 
%
%
On the other hand, we generate the HR data by decimating the original frames straight to $32\times32$ pixels. 
We perform \textit{leave-person-out} cross validation in each case and
compute correct classification rate (CCR) and standard deviation
(StDev) to measure performance. 
%
 %

We also test our algorithm on the popular HMDB dataset \cite{kuehne2011hmdb} used for video activity recognition. 
The HMDB dataset consists of 6,849 videos divided into 51 action categories, each containing a minimum of 101 videos. 
In comparison to IXMAS, which was collected in a controlled environment, the HMDB dataset includes clips from movies and YouTube
%
%
videos, which are not limited in terms of illumination and camera position variations.
Therefore, HMDB is a far more challenging dataset, especially when we decimate to eLR, which we herein refer to as eLR-HMDB.
In our experiments, we used the three training-testing splits provided with this dataset.
Note that since there are 51 classes
in the HMDB dataset, the CCR based on a purely random guess is
$1.96\%$.

\subsection{Results for eLR-IXMAS}
We first conduct a detailed evaluation of the proposed paradigms on
the eLR-IXMAS action dataset.
For a fair comparison, we follow the image resolution,
pre-processing and cross-validation as described in
\cite{dai2015towards}.
We first resize all video clips to a fixed temporal length $T = 100$ using cubic-spline interpolation.
%
%
%
%
%
%

Table~\ref{tbl:cc_results} summarizes the action recognition accuracy on the eLR-IXMAS dataset.
We report the CCR for separate spatial and temporal ConvNets, as well as for various locations and operators of fusion, with and without eLR-HR coupling. 
%
We also report the baseline result from \cite{dai2015towards} which employs a nearest-neighbor classifier. 

%
%
{\renewcommand{\arraystretch}{1.5} 
\begin{table}[thb]\footnotesize
	\vglue -0.0cm
	\caption{Performance of different ConvNet architectures against baseline on the eLR-IXMAS dataset. ``Spatial \& Temp avg'' has been performed by averaging the temporal and spatial stream predictions. The best performing method is highlighted in bold.}
	\label{tbl:cc_results}
	\centering
	{\footnotesize
		\begin{tabular}{|*{5}{c|}}
			
			 \hline
			 \multirow{2}{*}{Method}	
			 &  \multirow{2}{*}{Fusion Layer} 
			 &  \multirow{2}{*}{eLR-HR} 
			 &  \multirow{2}{*}{CCR} 
			 &  \multirow{2}{*}{StDev}
			 \\
			 &   & coupling?  &  &    \\
			 \hline
			 
             Baseline (Dai \cite{dai2015towards})  & -  & - & $ 80.0\%$      & $ 6.9\%$   \\
            
        	\hline
        	Spatial Network    	& - & No & $88.6 \% $    & $ 6.2\%$  \\
        	\hline
        	    
        	Temporal Network  	& - & No & $91.6 \% $  & $ 4.9\%$  \\
        	\hline
            Spatial$\&$Temp avg & Softmax & No & $92.0 \% $  & $ 6.0\% $ \\
            \hline
           
            \multirow{4}{*}{Concat Fusion} 	
            & Fc 4      & No & $ 92.2\%$      & $ 5.2\%$    \\  
            \cline{2-5}	
            & Fc 4    & Yes  & $ 92.5\%$   & $ 5.5\%$ \\
            \cline{2-5}		     		
            &  Conv 3  & No & $ 92.2\%$      & $ 5.2\%$\\
            \cline{2-5}	
            &  Conv 3  & Yes & $ 93.3\%$        & $ 5.6\%$\\
            \hline	
            
            \multirow{4}{*} {Conv Fusion} 	
            & Fc 4      & No &$ 92.0\%$      & $ 5.8\%$   \\ 
            \cline{2-5}	
            & Fc 4    & Yes  & $ 93.1\%$   & $ 5.2\%$ \\
            \cline{2-5}		     		
            & Conv 3   & No &$ {93.3\%}$      & $ {4.0\%}$   \\
            \cline{2-5}	
            & {\bf Conv 3}   & {\bf Yes} &$ \mathbf{93.7\%}$      & $ \mathbf{4.5\%}$   \\
            \hline	
      
           \multirow{4}{*}{Sum Fusion}	
           & Fc 4    & No  & $ 92.2\%$   & $ 5.5\%$ \\
           \cline{2-5}
           & Fc 4    & Yes  & $ 92.8\%$   & $ 7.1\%$ \\
           \cline{2-5}		     		
           & Conv 3 & No  & $ 93.0\%$   & $ 4.7\%$ \\
           \cline{2-5}	
           & Conv 3 & Yes  & $ 93.6\%$   & $ 4.0\%$ \\
           \hline
           
		\end{tabular}}
		\vglue -0.2cm
	\end{table}
}
\begin{figure}
	\centering
	\begin{subfigure}[b]{0.48\textwidth}
		\frame{\includegraphics[width=1\linewidth]{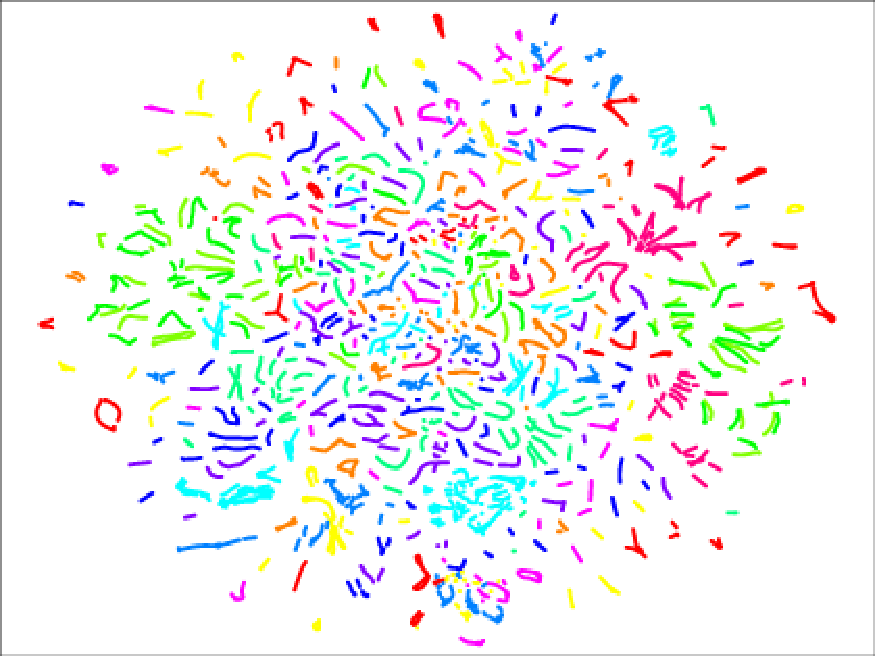}}
		\caption{eLR-IXMAS pixel-wise time series features \cite{dai2015towards}}
		\label{fig:Ng1} 
	\end{subfigure}
	
	\begin{subfigure}[b]{0.48\textwidth}
		\frame{\includegraphics[width=1\linewidth]{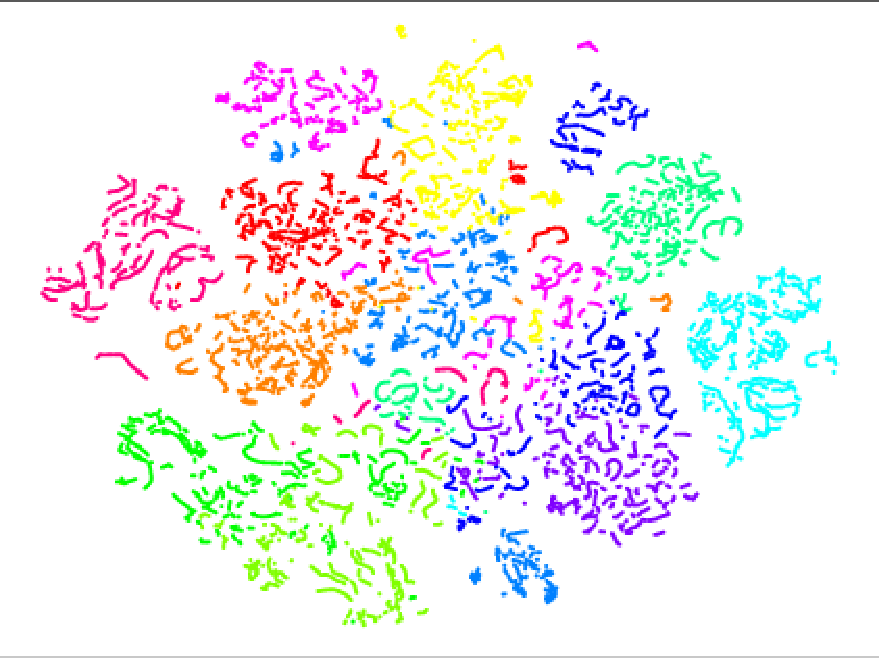}}
		\caption{eLR-IXMAS ConvNet features after `Fc 5' layer}
		\label{fig:Ng2}
	\end{subfigure}
	\caption{2-D t-SNE embeddings \cite{maaten2008visualizing} of
          features for the eLR-IXMAS dataset. A single marker
          represents a single video clip and is color-coded by action
          type. (a) Embeddings of pixel-wise time series features
          \cite{dai2015towards}. (b) Embeddings of the last
          fully-connected layer's output from our best performing
          ConvNet.}
	\label{fig:t-sne-ixmas}
\end{figure}
We first observe that dedicated spatial or temporal ConvNet outperforms 
the benchmark result from \cite{dai2015towards} by $8.6\%$ and $11.6\%$ respectively, which validates the discriminative power of a ConvNet. 
If we equally weigh these two streams (``Spatial\&Temp avg''), we can see that the fusion only marginally improves recognition performance. 
%
%
%
Secondly, we can see that fusing after the ``Conv3'' layer provides a consistently better performance than fusing after the ``Fc4'' layer. 
In our preliminary experiments, we also found that fusing after the ``Conv3'' layer was consistently better than fusing after the ``Conv2'' layer, which suggests that there is an ideal depth (which is not too shallow or too deep in the network) for fusion.
%
%
Regarding which fusion operator to use, we note that all 3 operators we consider provide comparable performance after the ``Fc4'' layer. 
However, if we fuse after the ``Conv3'' layer, convolutional fusion performs best.
%

As for the effectiveness of semi-coupling in the networks using HR information, we can see that eLR-HR coupling consistently improves recognition performance.
%
%
Our best result on IXMAS is $93.7\%$, where we find that without coupling, our performance drops by $0.4\%$.
This result is very close to that achieved by using {\it only} HR data in both training and testing, which is $94.4\%$ CCR.
Effectively, this should be an upper-bound, in terms of performance, when using eLR-HR coupling in training but testing {\it only} on eLR data.
%
%
%
That the performance gap between HR and eLR is small may be explained
by the distinctiveness of actions and the controlled indoor
environment (static cameras, constant illumination, etc.) in the IXMAS
dataset.
%
Additionally, the fine details (e.g., hair, facial features), that are
only visible in HR, are not critical for action recognition.

%

In order to qualitatively evaluate our proposed model, we visualize various feature embeddings for the eLR-IXMAS dataset. 
We extract output features of the ``Fc5'' layer from the best-performing ConvNet (shown in bold), and project them to 2-dimensional space using t-SNE \cite{maaten2008visualizing}.
%
For comparison, we also apply t-SNE to the pixel-wise time series features proposed in our benchmark \cite{dai2015towards}.
As seen in Fig.~\ref{fig:t-sne-ixmas}, the feature embedding from our ConvNet model is visually more separable than that of our baseline.
This is not surprising, as we are able to consistently outperform the baseline on the eLR-IXMAS dataset.
%

\begin{table}
	\vglue -0.0cm
	\caption{Comparison of the number of parameters of our best-performing action recognition ConvNet as compared to those of the standard-resolution image classification ConvNets.}
	\label{tbl:param}
	\begin{center}
		\begin{tabular}{|l|c|c|c|}
			\hline
%
%
			Network & Task & Input resolution & $\#$ param \\
			\hline\hline
			Ours       & Action Rec.   & $32\times32\times3$    & 0.84M\\
			AlexNet   & Image Class. & $224\times224\times3$ &  60M  \\
			VGG-16   & Image Class. & $224\times224\times3$ & 138M \\
			VGG-19   & Image Class. & $224\times224\times3$ & 144M \\
			\hline
		\end{tabular}
		\vglue -0.2cm
	\end{center}
\end{table}
Regarding the number of parameters, our ConvNet designed for eLR videos needs about 100 times less parameters than state-of-the art ConvNets designed for image classification like AlexNet \cite{krizhevsky2012imagenet},  VGG-16, and VGG-19 \cite{Simonyan14c} (Table~\ref{tbl:param}). 
%
%
In consequence, this significantly reduces the computation cost of training and testing compared to these standard-resolution networks.
\subsection{Results for eLR-HMDB}
%
 %
We also report the results of our methods on eLR-HMDB.
%
Note that, for this dataset, we only report results for fusion after the ``Conv3'' layer, based on our observations from eLR-IXMAS.
%
We follow the same pre-processing procedure as used for eLR-IXMAS except that we do not resize the video clips temporally for the purpose of having a fair comparison with the results reported in \cite{ryoo2016privacy}.
Our reported CCR is an average across the three training-testing splits provided with this dataset. 

First, we measure the performance of a dedicated spatial-stream ConvNet and a dedicated temporal-stream ConvNet. 
As shown in Table~\ref{tbl:HMDB_results}, using only the appearance information (spatial stream) provides $19.1\%$ accuracy.
If optical flow is used alone (temporal stream), performance drops to $18.3\%$.
%
This is likely because videos in the HMDB dataset are unconstrained; camera movement is not guaranteed to be well-behaved, thus resulting in drastically different optical-flow quality across videos.
%
Such variations are likely to be amplified in eLR videos. 
We then evaluate the same three fusion operators after the ``Conv3'' layer.
%
Not surprisingly, compared to the average of predictions from a
dedicated spatial network and a dedicated temporal network, fusing the
temporal and spatial streams improves the recognition performance by
$0.8\%$, $0.9\%$ and $1.8\%$ with concatenation, convolution, and sum
fusion, respectively.
Fusion alone does not bring significant improvement. This, however, is consistent with the observations in \cite{feichtenhofer2016convolutional}.
%
{\renewcommand{\arraystretch}{1.5} 
	\begin{table}[thb]\footnotesize
		\vglue -0.0cm
		\caption{Performance of different ConvNet architectures and current state-of-the-art method on the eLR-HMDB dataset. The two-stream networks are all fused after the ``Conv3'' layer. The best method is highlighted in bold.}
		\label{tbl:HMDB_results}
		\centering
		{\footnotesize
			\begin{tabular}{|*{3}{c|}}
				
				\hline
				\multirow{2}{*}{Method}	
				&  \multirow{2}{*}{eLR-HR} 
				&  \multirow{2}{*}{CCR} 
				\\
				&    coupling?  &  \\
				\hline
				
				Spatial Network    	& No & $19.1 \%$     \\
				\hline
				
				Temporal Network   & No & $18.3 \%$  \\
				\hline
				
				Spatial $\&$ Temp avg & No &  $19.6 \%$\\
				\hline
				
				\multirow{2}{*}{Concat Fusion} 	     		
				& No & $ 20.4\%$\\
				\cline{2-3}	
				& Yes & $ 27.1\%$\\
				\hline	
				
				\multirow{2}{*} {Conv Fusion} 	     		
				& No &$ {20.5\%}$        \\
				\cline{2-3}	
				& Yes & $ 27.3\% $  \\
				\hline	
				
		    	\multirow{2}{*}{Sum Fusion}	     		
		    	& No  & $  21.4\% $\\
		    	\cline{2-3}	
		    	& Yes  & $ \mathbf{29.2}\%$\\
		    	\hline		
		    	
		    	\multirow{2}{*} {ConvNet feat} 
		    	& 	\multirow{2}{*} {-} & 	\multirow{2}{*}{$18.9\%$} \\
		    	+  SVM\cite{ryoo2016privacy} & & \\
				\cline{1-3}
				
					\multirow{2}{*} {ConvNet feat} 
					& 	\multirow{2}{*} {-} & 	\multirow{2}{*}{$20.8\%$} \\
					+  ISR + SVM\cite{ryoo2016privacy} & & \\
					\cline{1-3}
					
				  	\multirow{2}{*} {ConvNet + hand-crafted feat} 
					& 	\multirow{2}{*} {-} & 	\multirow{2}{*}{$28.7\%$} \\
					+  ISR + SVM\cite{ryoo2016privacy} & & \\
					\cline{1-3}

				\hline
			\end{tabular}}
			\vglue -0.2cm
		\end{table}
	}
	
When fusion is combined with eLR-HR coupling, the gains are significant.
%
%
We achieve large performance gains from $20.4\%$ to $27.1\%$ using
concatenation fusion, $20.5\%$ to $27.3\%$ using convolutional fusion, and
$21.4\%$ to $29.2\%$ using sum fusion.
Such notable improvements validate the discriminative capabilities of
semi-coupled fused two-stream ConvNets.
%
Compared to the state-of-the-art results reported in
\cite{ryoo2016privacy}, our approach is able to outperform their
ConvNet feature-only method by $8.4\%$.
We also exceed the performance of their best method, that uses an
augmented hand-crafted feature vector, by $0.5\%$.
\section{Conclusion}
In this paper, we proposed multiple, end-to-end ConvNets for action
recognition from extremely low resolution videos (e.g., $16\times12$
pixels).
We proposed multiple eLR ConvNet architectures, each leveraging and fusing spatial and temporal information.
%
%
Further, in order to leverage HR videos in training we incorporated eLR-HR coupling to learn an intelligent mapping between the eLR and HR feature spaces.
The effectiveness of this architecture has been validated on two datasets.
We outperformed state-of-the-art methods on both eLR-IXMAS
and eLR-HMDB datasets.

\section{Acknowledgement}
We gratefully acknowledge the support of NVIDIA Corporation with the donation of the Titan X Pascal GPU used for this research.

%
%
%
%
%

{\small
\bibliographystyle{ieee}
\bibliography{egbib}
}

\end{document}